\pdfoutput=1
\documentclass[11pt]{article}

\usepackage[final]{acl}

\usepackage{times}
\usepackage{latexsym}
\usepackage{booktabs}
\usepackage{multirow}
\usepackage{colortbl}
\usepackage[T1]{fontenc}
\usepackage[utf8]{inputenc}

\usepackage{microtype}

\usepackage{inconsolata}

\usepackage{graphicx}
\usepackage{amsmath}
\newcommand{\topthree}[1]{#1{\color[HTML]{CB4335}$\dagger$}}

\newif\ifnotanonymize
\notanonymizetrue

\title{Mind the Gap! \\ Static and Interactive Evaluations of Large Audio Models}

\author{Minzhi Li$^{* \nu \sigma \S}$ \quad William Held\thanks {Equal Contribution. Work completed at Stanford University. Contacts: li.minzhi@u.nus.edu, held@stanford.edu, diyiy@stanford.edu.}$^{\gamma \sigma}$ \quad \textbf{Michael J. Ryan}$^{\sigma}$\\ \textbf{Kunat Pipatanakul}$^{\omega}$ \quad \textbf{Potsawee Manakul}$^{\omega}$  \quad \textbf{Hao Zhu}$^{\sigma}$  \quad \textbf{Diyi Yang}$^{\sigma}$ \\ $^{\gamma}$Georgia Institute of Technology $^{\nu}$National University of Singapore  \\ $^{\omega}$SCB 10X, SCBX Group 
$^{\S}$Institute for Infocomm Research (I$^2$R), A*STAR \\ $^{\sigma}$Stanford University\\
\href{https://talkarena.org}{TalkArena.org}
}

\begin{document}
\maketitle
\begin{abstract} 
As AI chatbots become ubiquitous, voice interaction presents a compelling way to enable rapid, high-bandwidth communication for both semantic and social signals. This has driven research into Large Audio Models (LAMs) to power voice-native experiences. However, aligning LAM development with user goals requires a clear understanding of user needs and preferences to establish reliable progress metrics. This study addresses these challenges by introducing an interactive approach to evaluate LAMs and collecting 7,500 LAM interactions from 484 participants. Through topic modeling of user queries, we identify primary use cases for audio interfaces. We then analyze user preference rankings and qualitative feedback to determine which models best align with user needs. Finally, we evaluate how static benchmarks predict interactive performance - our analysis reveals no individual benchmark strongly correlates with interactive results ($\tau \leq 0.33$ for all benchmarks). While combining multiple coarse-grained features yields modest predictive power ($R^2$=$0.30$), only two out of twenty datasets on spoken question answering and age prediction show significantly positive correlations. This suggests a clear need to develop LAM evaluations that better correlate with user preferences.

\end{abstract}

\section{Introduction}
Compared to text, speech enables faster, more efficient interaction~\citep{ruan2016speech} and further enables communication of paralinguistic information~\citep{sutton2019voice}. These dual motivations make speech interaction a promising step towards more ubiquitous computing~\citep{ubiq_jl}. Following this vision, researchers have developed large language models (LLMs) that directly accept audio inputs~\citep{latif2023sparks, tang2023salmonn, chu2024qwen2, chu2023qwen, hurst2024gpt, held2024distilling}, known as Large Audio Models (LAMs)~\citep{latif2023sparks}.

Recent work has focused on evaluating these systems by aggregating existing static benchmarks~\citep{yang2024air, wang2024audiobench}. Similar to traditional LLM evaluation approaches~\citep{few_shot}, these frameworks examine LAMs in zero-shot and few-shot settings using tasks originally designed for finetuned models. The evaluations cover both speech interaction capabilities where natural text counterparts exist~\citep{wu2023heysquad, wang2020covost} and paralinguistic feature recognition~\citep{busso2008iemocap, hasan-etal-2019-ur}, reflecting the dual motivations driving LAM development.

\begin{figure}[t!]
    \centering \includegraphics[width=\linewidth]{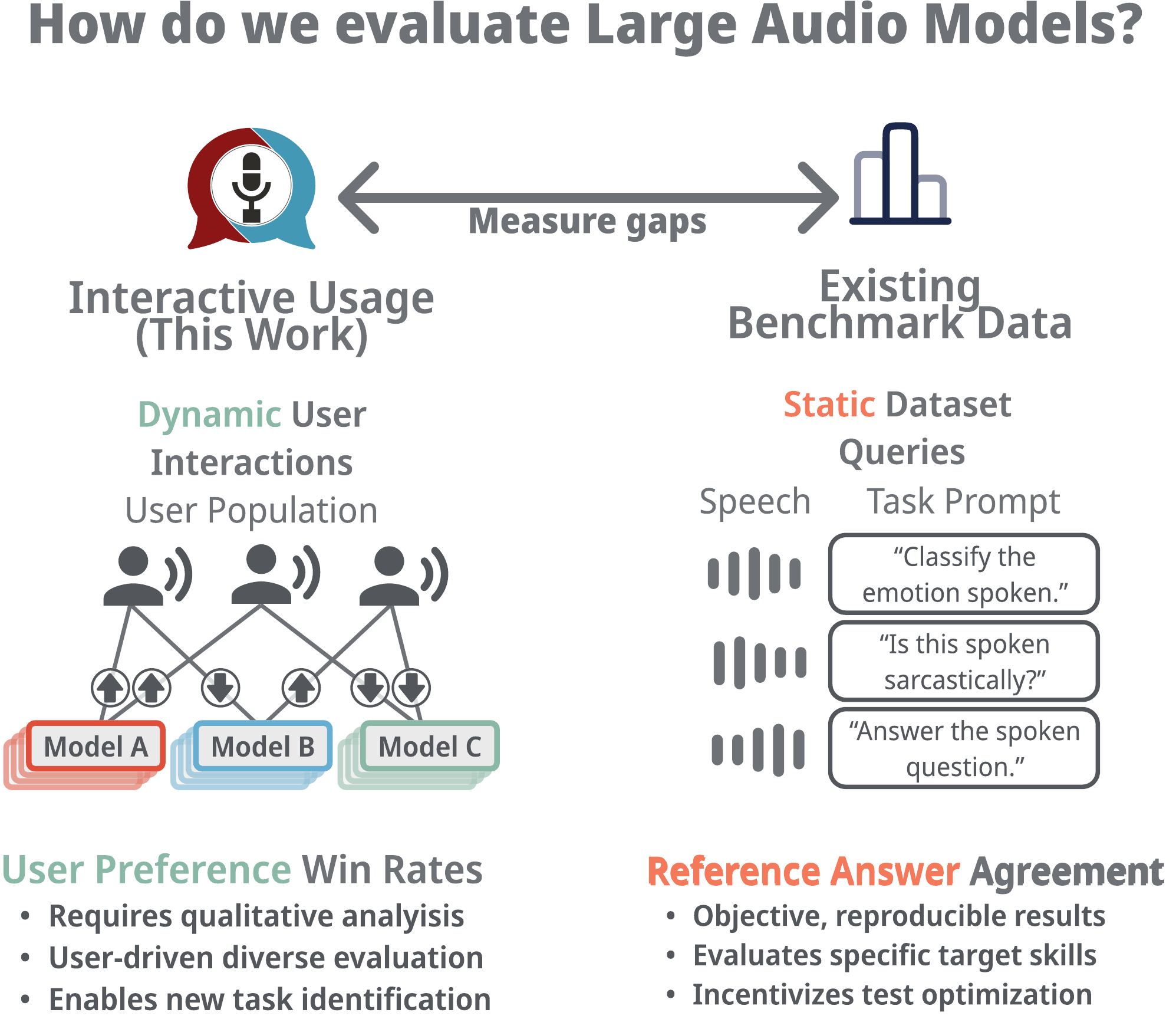}
    \caption{Comparison of static and interactive ways of evaluating Large Audio Models. In this work, we perform interactive evaluations to understand how LAMs are likely to be used and how they can be benchmarked.} 
    \label{fig:crown_jewel}
\end{figure}

While static benchmarks are ideal for measuring progress on specific capabilities, their value diminishes when gaps exist between measured capabilities and user expectations~\citep{lee2022evaluating}. For this reason, benchmarks for text-only LLMs typically aim to correlate with real-world user preferences captured through interactive evaluations like Chatbot Arena~\citep{chiang2024chatbot}. Indeed, popular text-only benchmarks such as ARC-C~\citep{clark2018thinksolvedquestionanswering}, MMLU~\citep{hendrycks2021measuringmassivemultitasklanguage}, and AlpacaEval~\citep{dubois2024alpacafarmsimulationframeworkmethods} all show high correlation ($\rho > 0.8$) with interactive user evaluation results~\citep{dubois2024lengthcontrolledalpacaevalsimpleway}.

However, it remains unclear whether existing LAM benchmarks correlate similarly with user evaluation, as no prior work has collected LAM preference data. This knowledge gap leaves LAM researchers without clear direction for model development that better align with user needs.
In light of this, we collect 7,500 interactions between 484 paid participants (all with prior commercial LLM chat experience) and 6 speech-in text-out systems, including 5 top-performing LAMs and a baseline system combining ASR with a text-only LLM. This preference data allows us to address questions beyond the scope of existing benchmark-based or text-only interactive evaluations:

\begin{itemize}
\setlength\itemsep{0em}
\item \textbf{What tasks do users expect LAMs to perform?} Our interface provides users with a simple audio interface and prompts them to test capabilities they would expect from an AI voice assistant. Following \citet{tamkin2024clioprivacypreservinginsightsrealworld}, we use LLMs to cluster and categorize these queries. We find that 77\% of usage falls into categories where speech primarily serves efficiency purposes (e.g., task execution) rather than conveying information unique to the audio modality.

\item \textbf{What models are best at these tasks and why?} Users receive text responses from randomly sampled systems and provide pairwise preferences with qualitative feedback. Surprisingly, we find that a pipeline system that combines  Whisper~\citep{radford2022whisper} and Llama~\citep{llama3modelcard} is preferred the most. This likely stems from two factors: most user queries rely primarily on text semantics, and three of the five most common feedback types focus on style of the textual output.

\item \textbf{Which benchmarks are the best proxies for user preferences?}
Our evaluation of LAMs across twenty existing speech benchmarks reveals limited predictive power for user preferences. No single benchmark shows strong correlation with human evaluations, and even aggregated benchmarks explain only 30\% of preference variance ($R^2 = 0.30$). Only two metrics show positive correlation: speech comprehension ability (measured via Public-SG-Speech) and reduced systematic errors (captured by an age prediction task where all evaluated models perform worse than random chance). This starkly contrasts text-only LLMs, where static evaluations and interactive assessments show high correlation, suggesting the need for new static audio evaluations, which our interaction data can inform.
\end{itemize}

\section{Related Work}
\paragraph{Large Audio Models.} Large-scale self-supervised audio models have been used to learn generalized audio representations from extensive unlabeled datasets. Early successful approaches such as wav2vec~\citep{schneider2019wav2vec} and  HuBERT~\citep{hsu2021hubert} learned audio representations from scratch, achieving robust performance across many tasks when finetuned. Focused primarily on scaling data and training time, recent efforts such as Whisper~\citep{radford2022whisper} and OWSM~\citep{owsm} have led to extremely effective models both for transcription and speech understanding.

Recent advancements in audio models have integrated learned audio representations with text-based LLMs, enabling native audio understanding while leveraging knowledge and stylistic insights from textual resources. This has led to the emergence of Large Audio Models (LAMs)~\citep{latif2023sparks}. Such models include SpeechGPT \citep{zhang2023speechgpt} which leverages HuBERT \citep{hsu2021hubert} for extracting continuous speech as discrete units, LLaMA \citep{touvron2023llama} as the text-LLM foundation, and HiFi-GAN \citep{kong2020hifi} as the unit vocoder; LTU \citep{gong2023listen} which consists of an audio spectrogram transformer, LLaMA and a Low-rank Adapter; Qwen-Audio series \citep{chu2023qwen, chu2024qwen2} with Whisper-large-v2 and Whisper-large-v3 as the audio encoder and Qwen-7B as the LLM, and many other Large Audio Models \citep{borsos2023audiolm, liu2023audioldm, held2024distilling}. In our work, we evaluated nine different LAMs that are publicly available on static benchmarks and tested five best-performing ones in the interactive setting.

\paragraph{Evaluation of Large Audio Models.}
To evaluate the audio processing capability of different models, prior research has constructed a variety of audio benchmarks, targeting particular abilities. For automatic speech recognition, benchmarks such as Librispeech \citep{panayotov2015librispeech} and Commonvoice \citep{ardila2019common} are widely used, with metrics like word error rate (WER) and character error rate (CER). For speech translation tasks, there are datasets like Covost \citep{wang2020covost}, Covost2 \citep{wang2021covost}, and CVSS \citep{jia2022cvss} with evaluation metrics such as BLEU scores. For emotion detection, benchmarks include MELD \citep{poria2018meld} and IEMOCAP \citep{busso2008iemocap} with speech data labeled with different emotions. In the domain of Speech Question Answering, there are SDQA \citep{faisal2021sd}, Social IQ 2.0 \citep{siq2}, and HeySquad \citep{wu2023heysquad}.

However, one problem regarding the evaluation of LAMs is that they have reported evaluation results on different sets of benchmarks, resulting in inconsistent evaluation and difficulty in comparison \citep{wang2024audiobench}. Therefore, there are commendable efforts to aggregate audio datasets together to evaluate LAMs in a holistic way such as AIRBench \citep{yang2024air}, AudioBench \citep{wang2024audiobench}, and VoiceBench \citep{chen2024voicebench}. However, they still utilize static reference-based metrics like WER and accuracy. In contrast, we interactively evaluate LAMs using user preferences.

\paragraph{Interactive Evaluation of LLMs}
Interactive evaluation can overcome many limitations in using static datasets to evaluate models. One limitation is model overfitting \citep{ying2019overview} where models are over-optimized for specific datasets and tasks, limiting their generalization capability. Moreover, static benchmarks may have data contamination \citep{magar2022data} issues where LLMs have been trained on the data. Furthermore, static evaluation may lack the ability to incorporate real-world scenarios \citep{lin2024wildbench} and align with human preferences \citep{oren2023proving}. Moreover, data drift \citep{mallick2022matchmaker} can happen when the environment generating the data evolves, causing a mismatch between the static datasets and the data in real-world scenarios. Thus, static datasets can fail to keep track of long-term model performance over time. These limitations strongly suggest the need for interactive evaluation of models.

As such, there are many research efforts on creating live NLP benchmarks. For example, DynaBench \citep{kiela2021dynabench} builds an open platform for dynamic data curation, and Chatbot Arena \citep{chiang2024chatbot} benchmarks models through chat with LLMs from a larger user base. In a similar line, there are works extending to other modalities and use cases like Wildvision-Arena \citep{lu2024wildvision} for vision-language models, Long Code Arena \citep{bogomolov2024long} for coding, and Web-Arena \citep{zhou2023webarena} for web-related tasks. To our knowledge, there is no similar interactive evaluation of audio-language models to investigate the gap between static benchmarks and user interactions.

\section{Interactive Evaluation}
To capture real-world use cases of LAMs, we collect user preferences on an open platform\ifnotanonymize\footnote{\url{https://talkarena.org/}}\fi. We then convert pairwise votes to model ranks that reflect the interaction capability of different models.
\subsection{Interface}

Our interface (see Figure \ref{fig:interface} in Appendix) is built using the Gradio platform~\citep{abid2019gradio}. This allows us to serve a simple web-based user experience with integration to both locally hosted and API-accessible LAMs.

\paragraph{User Interaction and Input}
Upon arriving on the platform, users are instructed to interact with the model for any use cases they would expect from a voice-based AI assistant. By providing no concrete example tasks, our goal was to capture diverse desired use cases with minimal bias based on our preconceptions of "interesting" or "challenging" use cases. Each query is streamed to the user character by character to avoid users being able to learn a mapping between tokenizations and specific model identities.

\paragraph{Pairwise Model Comparison.}
After submitting a query, users receive responses from two anonymous models, which are randomly selected and ordered in order to avoid personal and positional bias in their preferences. For assessment, users provide a simple \textbf{pairwise preference ranking}---choosing the better response or indicating no preference between the two. This method provides a relative ranking rather than an absolute performance score, allowing the user to make a simple decision and avoiding performance ceilings, which may be induced by reference-driven evaluation.

\paragraph{User Feedback and Justification.}
One shortcoming of preference ratings is that they offer minimal insights into the factors that drive user preferences. While this allows for very open-ended user values to be integrated into final model ratings, it also makes the data less directly valuable for deriving insights about what needs improvement in existing models.
To gain deeper insights, users can optionally justify their choices via text or speech, following prior work showing that users provide dramatically more detail when given a speech interface~\citep{molmo}. Even without requiring the completion of this field, we find that 44.9\% of users opted to provide qualitative rationale.

\subsection{Data Collection}
This research has been approved by the Institutional Review Board (IRB) at the authors' institution.  We selected Prolific as the survey platform due to its large participant pool and high average data quality \cite{eyal2021data, douglas2023data}. Since the average crowd worker is likely not a user of AI voice chat products, we further used the platform's pre-screening to select participants who reported actively having used LLM chatbots such as ChatGPT and Gemini previously. The only other limiting requirement was that participants had access to a working microphone to record and submit their voice.

For each pair of models we evaluated, we recruited 50 participants, and each participant can contribute 10 votes. This allows us to obtain a sufficient pool of votes that can illustrate user preference between model pairs in a statistically significant manner. In total, 7500 votes were collected from a diverse pool of around 484 unique participants. We also apply the selection criteria to ensure the pool of participants is gender-balanced to allow a fair representation of user preference. Each user was paid \$2.50 for 10 votes, with a minimum of \$15 per hour ensured.

\subsection{Model Rank}
To convert the collected pairwise preference data to model ranking, we apply the Bradley-Terry model \citep{bradley1952rank} to compute scores for each model for its statistical rigor and good interpretability. The Bradley-Terry model provides a principled way to infer latent winning ability to estimate the probability of one entity winning over another. The model defines an exponential score function $p_i$ as $e^{\beta_i}$ for model $i$ where $\beta$ is the Bradley-Terry coefficients. For a model pair of model $i$ and model $j$, the probability of model $i$ being preferred over model $j$ is given by Equation \eqref{eq:probability}:

\begin{equation}
    \label{eq:probability}
    \Pr(i > j) = \frac{e^{\beta_i}}{e^{\beta_i} + e^{\beta_j}}
\end{equation}

The Bradley-Terry coefficients $\beta$ are computed by maximizing the log-likelihood of the observed pairwise preferences $\mathcal{D}$, given by 

\begin{align}
  \mathcal{L}(\beta) = \sum_{(i,j,y) \in \mathcal{D}} &[y \log(\sigma(\beta_i - \beta_j))  +\nonumber \\ 
  &(1-y)\log(\sigma(\beta_j - \beta_i))]
\end{align}

where $y$ is 1 when $i$ is preferred over $j$, 0.5 for ties, and 0 when $j$ is preferred; $\sigma$ is the sigmoid function. The optimization is performed using the L-BFGS algorithm\footnote{We follow the updated methodology used by \href{https://www.claytonthorrez.com/blog/posts/fast_llm_ratings/}{Chatbot Arena} without scaling or normalization.}. We then compute $p_i = e^{\beta_i}$.

With quality scores $\mathbf{p}$ estimated, the models can be ranked in a descending order based on the scores. A higher score indicates a higher likelihood of being preferred by users.

\section{What Tasks Do Users Expect LAMs to Perform?}
\label{sec:clio}

\begin{figure}[t]
    \centering \includegraphics[width=\linewidth]{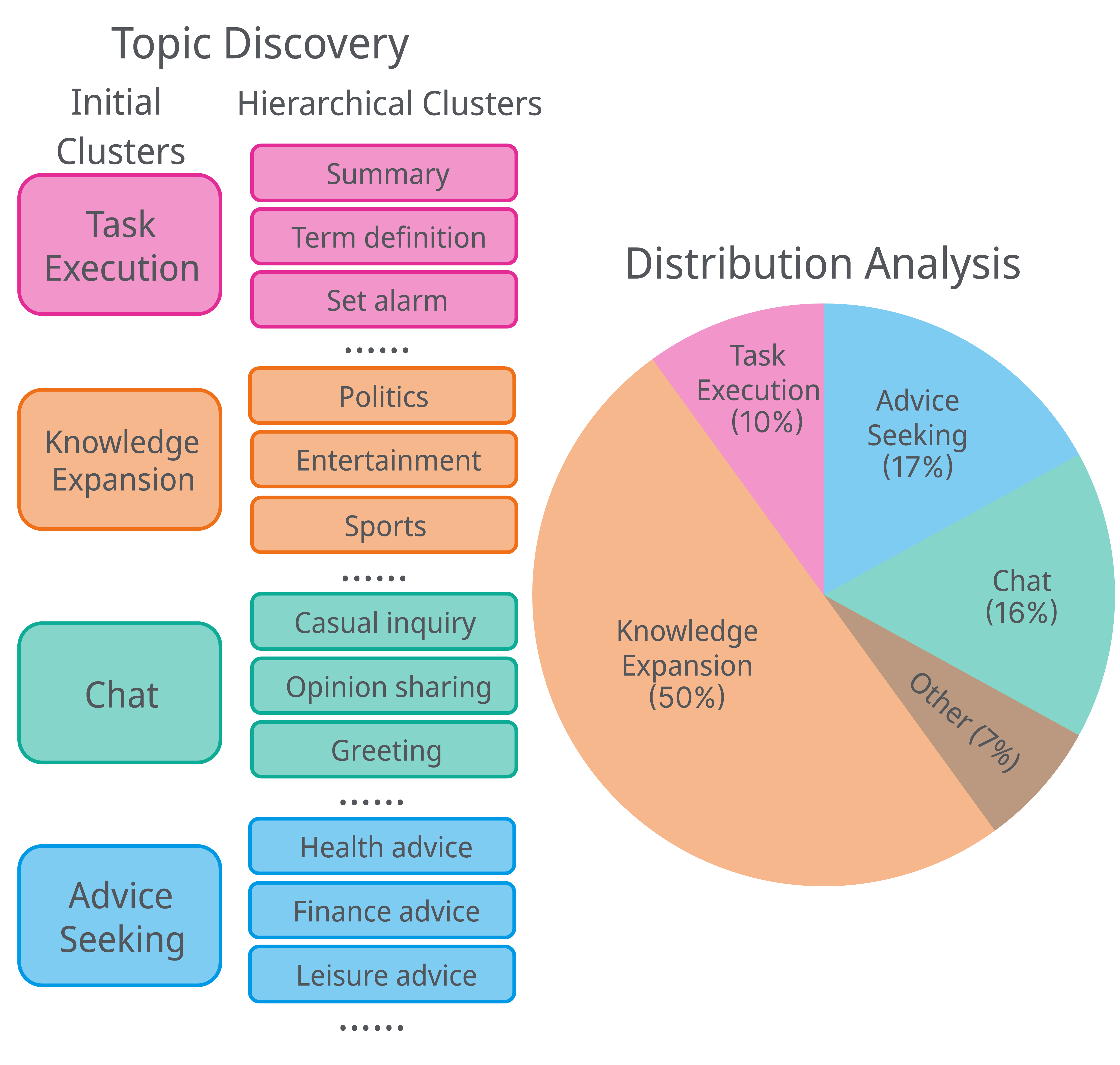}
    \caption{We identify four main topics in user queries --- task execution, knowledge expansion, chat, advice seeking as well as sub-topics under each category through hierarchical clustering (left) and analyze the relative proportions of each query type (right).} 
    \label{fig:user_query}
% \vspace{-0.3cm}
\end{figure}

We adopt the Clio analysis flow \citep{tamkin2024clioprivacypreservinginsightsrealworld} on transcripted user queries to explore topics in users' queries. We first apply the K-Means clustering algorithm on BERT embeddings of summaries of 1000 randomly sampled queries and identify \textbf{task execution}, \textbf{knowledge expansion}, \textbf{chat}, and \textbf{advice seeking} as four initial clusters through the Elbow Method and merging similar clusters after manual inspection. We then discover hierarchical clusters through recursive application of clustering algorithm as shown in Figure \ref{fig:user_query}.

To better understand the topic distribution, we manually listen to 100 randomly sampled recordings of user queries and classify them into one of the four main topics discovered. We found most users ask about knowledge-related questions (50\%) (e.g. \textit{``What is galaxy?''}), followed by advice seeking (17\%) (e.g. \textit{``I'm thinking of getting some brine shrimp. What should I know before I get them?''}), chat (16\%) (e.g. \textit{``Good morning, how are you?''}), and task execution (10\%) (e.g. \textit{``Summarize Volume 1 of Lord of the Mysteries.''}). These dominant uses suggest about areas LAMs could work towards to better user experience: it is important for LAMs to be equipped with up-to-date and comprehensive factual knowledge, potentially through methods like retrieval augmentation, to address the key use of knowledge query. Moreover, emotional and contextual understanding is crucial in advice seeking and chat situations. Furthermore, accurate intent detection is needed for task execution tasks. Moreover, 7\% of the sampled recordings include background noises such as music and voices of other people who is not the speaker. This introduces an additional factor beyond textual content to be considered in improving LAMs' interactive capability during real-world interactions.

Compared to top user queries during interactions with text-LLMs~\citep{zheng2023lmsyschat1m}, participants rarely talk about mathematical concepts and coding problems. This is expected as math and coding require precise syntax, symbols, and formatting, which can be difficult to dictate naturally. This suggests that when mapping text queries to audio inputs, we should consider distributional differences between written and spoken language for data curation.

\section{What Models are Best at these Tasks and Why?}

\paragraph{User Preferences}
\begin{figure*}[t]
\includegraphics[width=0.58\textwidth]{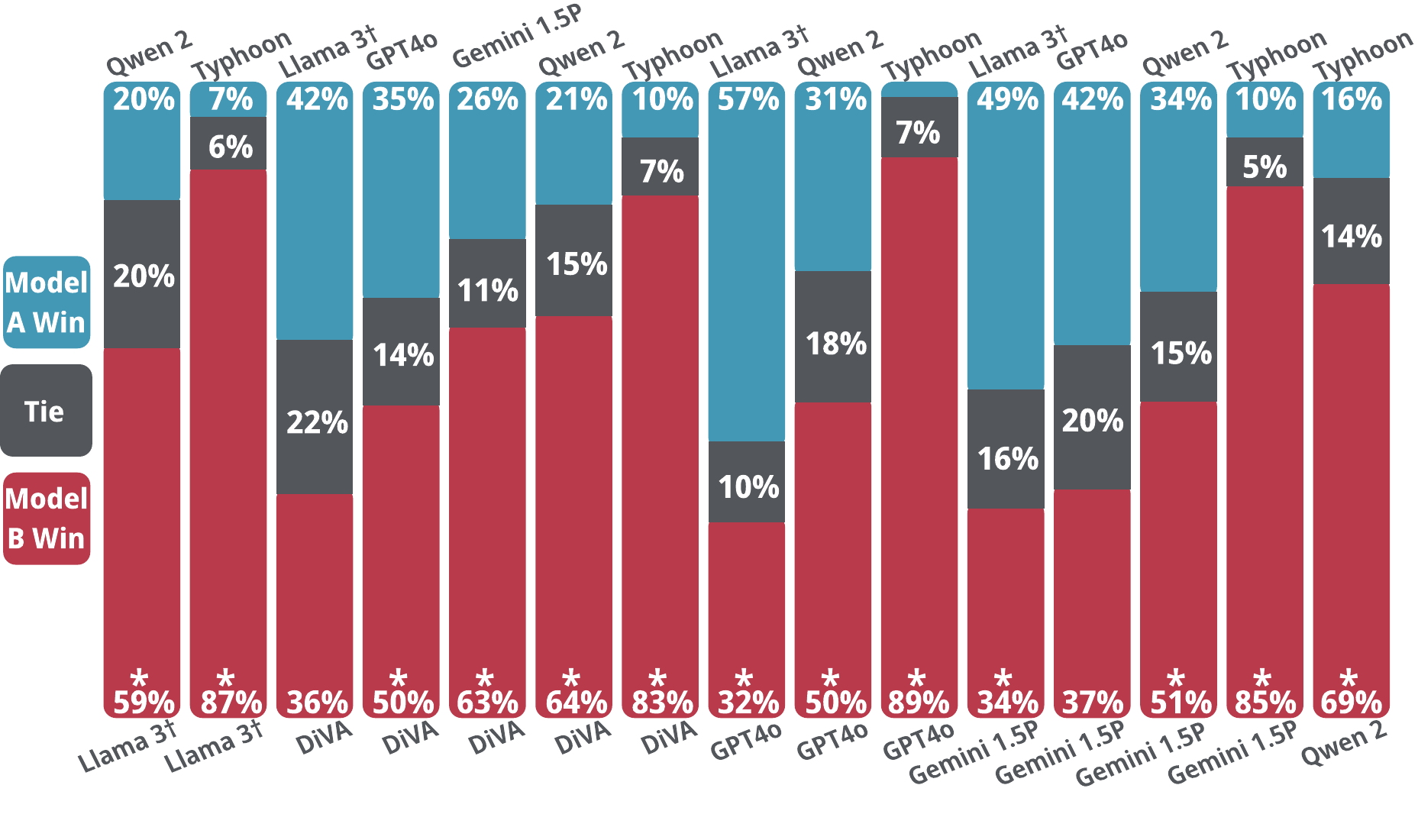}\hfill
\includegraphics[width=0.4\textwidth,trim=0 1in 0 0,clip]{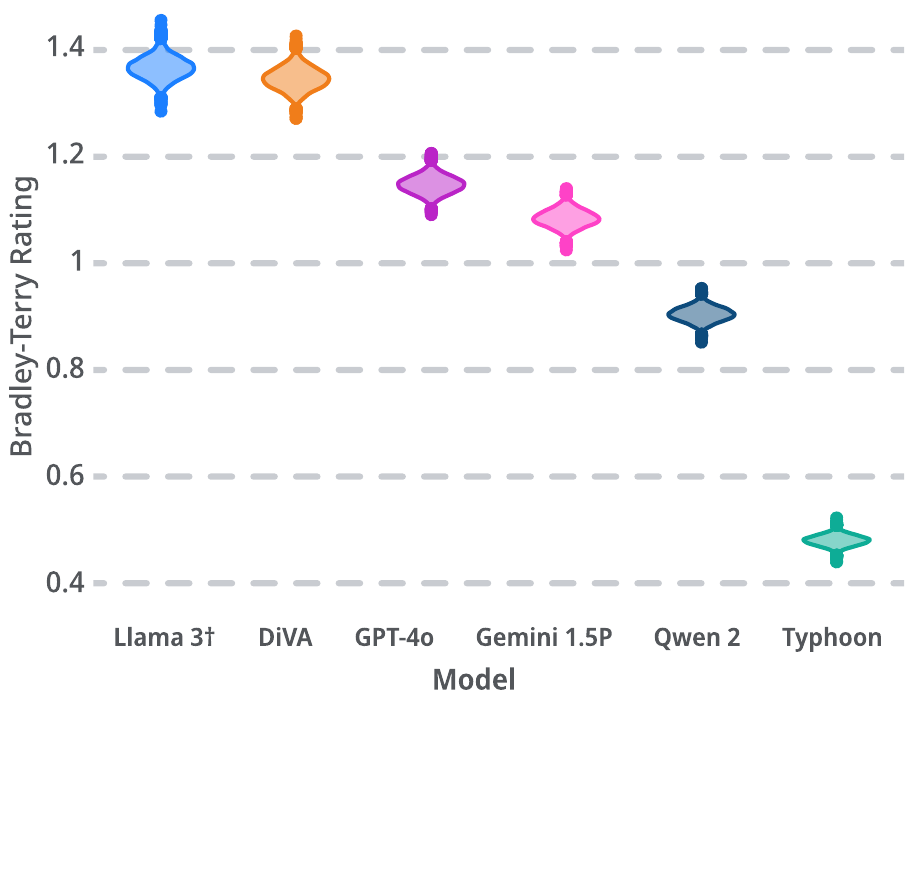}
  \caption{Head-to-head model comparisons (left) and Bradley-Terry (right) Scores from our evaluation. For win rates, * indicates the difference between preferences is significant (P<0.05) by a pairwise bootstrap test. For Bradley-Terry scores, distributions are shown shown for 10,000 bootstraps. $\dagger$ denotes an ASR + LLM pipeline.}
  \label{fig:votes}
\end{figure*}
In Figure \ref{fig:votes}, although close-sourced models like GPT4o demonstrates the best performance on static benchmarks, it is not the most preferred model in voice-in text-out interactions. Instead, the ASR pipeline of whisper-v2.0 and Llama3-8B-Instruct is most preferred among all six model settings, followed by DiVA which is trained by distilling a text LLM. This is because most of the user queries do not rely on nuanced speech understanding (Figure \ref{fig:user_query}) as we place minimal constraints on users' queries without asking them to submit challenging samples. It also suggests that the most effective way to improve current models' interactive capability for general single-turn use cases is to \textbf{leverage a powerful text language model's interactive capabilities}. 

\paragraph{Reasons for Preferences}
We manually analyze 100 randomly sampled user explanations for non-tie votes and summarize five most commonly observed reasons for users' preferences.
31\% of user explanations mention about \textbf{(1) level of details} in the text responses during interactions. In general, users find a model more preferable if they can generate \textit{specific, concrete and in-depth} responses (e.g. \textit{``I think both models here were able to answer me a bit, but certainly model two answered me more in-depth, so we're going to go with that one.''}). Another important factor in determining users' preferences is \textbf{(2) helpfulness} (24\%) of the response. Users will not prefer a model which refuses to respond to their questions or fail to address their questions (e.g. \textit{``Model 2 is completely irrelevant and refuses to answer the question I asked because it claims it is political, which it is not.''}). On the other hand, they prefer models which understand their needs and provide \textit{useful} feedback which \textit{adheres to their instructions}. 

Moreover, \textbf{(3) language appropriateness} (12\%) of the response can affect users' interaction experience with the models. Some models generated responses in a different language from the user query, and users find such responses \textit{illegible} and less preferred. This also observed through relatively similar model rank in interactive setting and model rank on the language detection benchmark. On top of that, users find \textbf{(4) accuracy} (11\%) of responses key. They prefer models which provide numerically and factually correct answers (e.g. \textit{``Model 1 is the factually accurate and most detailed and preferable response.''}). 

Furthermore, \textbf{(5) human-likeness} was mentioned in 11\% of user explanations sampled. Interestingly, some prefer responses that are not human-like (e.g. \textit{``I like the AI that admits it cannot have opinions.''}) while others prefer the model which possesses human traits (e.g. \textit{``Model1 was friendly and inquisitive''}). This shows the degree of human likeness is an important dimension that could affect interaction experience but needs adaptations to different users' preferences.

\section{Which Benchmarks are the Best Proxies for User Preferences?}
To understand the degree to which current speech benchmarks reflect the interaction capability of LAMs, we perform static evaluation on a comprehensive set of benchmarks that measure capabilities that may affect users' interaction experience. We run a logistic regression analysis to investigate the predictive power of each static dataset and obtain insights on directions future datasets should consider to align with real-world user experience.
\subsection{Dataset Selection}

We construct a superset of 20 datasets related to speech understanding and interactions from existing aggregated evaluation sets for LAMs (AudioBench \citep{wang2024audiobench} and AIR-Bench \citep{yang2024air}). The datasets cover a wide range of tasks that evaluate models' knowledge of \textbf{Speaker Cognitive State}, \textbf{Speaker Identity}, and \textbf{Speech Content Understanding}. 
Our goal is to evaluate a comprehensive set of tasks that potentially influence user experience during interactions. This set can then be filtered to identify the most predictive tasks in an unopinionated fashion.

\paragraph{Speaker Cognitive State} The ability to understand the cognitive states of speakers can be closely related to the interaction capability of models as effective interactions depend on accurate interpretation of intents and emotions \citep{tomasello2023having, jensen2016affect}. For intent detection, we include datasets on pragmatic intent detection \citep{bastianelli-etal-2020-slurp}, humorous intent detection \citep{hasan-etal-2019-ur}, and sarcastic intent detection \citep{castro-etal-2019-towards}. For emotion recognition, we include the IEMOCAP \citep{busso2008iemocap} and MELD \citep{poria2018meld} datasets.

\paragraph{Speaker Identity} Understanding speakers' identity is crucial in context-dependent and personalized interactions \citep{wu2024speaker}. We include audio benchmarks with annotated fields on speaker identity (language, accent, demographic information, social relationship). We evaluate models on tasks like language identification \citep{yang2024air}, accent classification \citep{ardila2019common}, gender and age classification \citep{ardila2019common, veliche2024measuringfairnessspeechrecognition}, as well as the relationship classification \citep{ardila2020commonvoicemassivelymultilingualspeech}.

\begin{table*}[t]
\centering
\newcommand{\numsize}[1]{{\huge #1}}
\def\arraystretch{2.0} % Adjusted line height
\resizebox{\textwidth}{!}{  \fontsize{16}{16}\selectfont
\begin{tabular}{
l 
c 
c 
c 
c 
c 
c 
c 
c 
c 
c 
c 
c 
c 
c }
\toprule
\textbf{Model} & \textbf{Humor} & \textbf{Sarcasm} & \textbf{Intent} & \textbf{Emotion} & \textbf{Relation} & \textbf{Gender} & \textbf{Age} & \textbf{Accent} & \textbf{Grounding} & \textbf{Language} & \textbf{Entity} & \textbf{QA} & \textbf{Instruction} & \textbf{ASR} \\ 
\midrule
\multicolumn{15}{l}{\textit{\fontsize{16}{16}\selectfont Commercial LAMs}} \\
\midrule
\textbf{GPT4o}  & \topthree{}\numsize{44.6} & \topthree{}\textbf{\numsize{53.6}} & \topthree{}\numsize{89.2} & \topthree{}\textbf{\numsize{29.1}} & \topthree{}\textbf{\numsize{59.7}} & \numsize{13.6} & \topthree{}\numsize{12.2} & \topthree{}\textbf{\numsize{35.3}} & \topthree{}\numsize{22.2} & \topthree{}\textbf{\numsize{73.3}} & \numsize{35.8} & \topthree{}\textbf{\numsize{65.2}} & \topthree{}\numsize{64.0} & \numsize{0.19} \\ 
\textbf{Gemini}  & \numsize{35.7} & \numsize{36.0} & \topthree{}\textbf{\numsize{91.4}} & \topthree{}\numsize{27.2} & \numsize{35.9} & \topthree{}\numsize{43.9} & \numsize{7.9} & \topthree{}\numsize{24.5} & \topthree{}\textbf{\numsize{25.9}} & \topthree{}\numsize{68.8} & \numsize{23.6} & \topthree{}\numsize{64.2} & \numsize{59.2} & \topthree{}\numsize{0.17} \\ 
\midrule
\multicolumn{15}{l}{\textit{\fontsize{16}{16}\selectfont Open-Weights LAMs}} \\
\midrule

\textbf{Qwen2-audio} & \numsize{34.9} & \topthree{}\numsize{41.5} & \topthree{}\numsize{81.1} & \numsize{23.2} & \numsize{17.3} & \topthree{}\textbf{\numsize{69.1}} & \topthree{}\textbf{\numsize{12.3}} & \numsize{5.4} & \numsize{10.0} & \topthree{}\numsize{66.5} & \topthree{}\textbf{\numsize{43.7}} & \topthree{}\numsize{62.3} & \numsize{62.6} & \topthree{}\numsize{0.16} \\
\textbf{Typhoon}  & \topthree{}\numsize{44.6} & \topthree{}\numsize{48.8} & \numsize{45.3} & \numsize{21.5} & \topthree{}\numsize{44.2} & \topthree{}\numsize{55.3} & \numsize{5.0} & \numsize{7.9} & \topthree{}\numsize{22.1} & \numsize{36.4} & \topthree{}\numsize{38.5} & \numsize{52.6} & \topthree{}\textbf{\numsize{68.3}} & \numsize{0.50} \\ 
\textbf{DiVA}  & \topthree{}\textbf{\numsize{46.2}} & \numsize{38.3} & \numsize{61.5} & \topthree{}\numsize{25.2} & \numsize{34.9} & \numsize{30.5} & \numsize{10.4} & \numsize{13.0} & \numsize{17.3} & \numsize{46.5} & \numsize{18.8} & \numsize{50.5} & \topthree{}\numsize{66.6} & \numsize{0.83} \\ 
\textbf{Qwen-audio} & \numsize{39.9} & \numsize{30.8} & \numsize{69.1} & \numsize{16.4} & \numsize{30.9} & \numsize{45.5} & \numsize{8.4} & \numsize{5.0} & \numsize{5.0} & \numsize{58.1} & \topthree{}\numsize{38.7} & \numsize{60.3} & \numsize{45.6} & \topthree{}\textbf{\numsize{0.07}} \\
\textbf{NExTGPT}  & \numsize{26.6} & \numsize{16.9} & \numsize{12.7} & \numsize{8.6} & \numsize{27.4} & \numsize{24.1} & \numsize{8.5} & \numsize{6.8} & \numsize{8.7} & \numsize{26.4} & \numsize{12.2} & \numsize{37.9} & \numsize{6.4} & \numsize{2.37} \\ 
\textbf{PandaGPT}  & \numsize{42.6} & \numsize{33.4} & \numsize{13.9} & \numsize{11.1} & \topthree{}\numsize{44.2} & \numsize{42.5} & \topthree{}\numsize{11.7} & \numsize{4.0} & \numsize{8.7} & \numsize{33.5} & \numsize{17.6} & \numsize{39.5} & \numsize{25.7} & \numsize{3.34} \\ 
\midrule
\multicolumn{15}{l}{\textit{\fontsize{16}{16}\selectfont Baselines}} \\
\midrule
\textbf{ASR Pipeline}  & \numsize{37.8} & \numsize{32.8} & \numsize{64.8} & \numsize{24.0} & \numsize{22.8} & \numsize{31.4} & \numsize{9.7} & \topthree{}\numsize{13.9} & \numsize{20.4} & \numsize{50.4} & \numsize{16.5} & \numsize{56.5} & \numsize{54.4} & \numsize{0.25} \\ 
\textbf{Random Baseline} & \numsize{50.0} & \numsize{50.0} & \numsize{25.0} & \numsize{25.0} & \numsize{25.0} & \numsize{50.0} & \numsize{14.3} & \numsize{20.0} & \numsize{25.0} & \numsize{25.0} & \numsize{25.0} & - & - & - \\ 
\bottomrule
\end{tabular}}

\caption{Average performance of LAMs and random baseline on 20 different benchmarks across 14 different speech understanding tasks. Top model performance is \textbf{bolded}, and top three model performances are marked with \topthree{}.} 
\label{tab:result}
\end{table*}

\paragraph{Speech Content Understanding} Another component fundamental to models' interaction capability is the ability to understand speech content \citep{greenberg1996understanding}. Besides traditional automatic speech recognition (ASR) tasks \citep{ardila2019common, panayotov2015librispeech}, it also involves understanding the entities, events, and user instructions in the speech. We evaluate models' ability for speech grounding \citep{wang2024audiobench}, speech entity recognition \citep{bastianelli-etal-2020-slurp}, speech instruction following \citep{wang2024audiobench}, and speech question answering \citep{wang2020covost2massivelymultilingual} tasks.

\subsection{Experiment Setup}
\paragraph{Models} We evaluate 9 different LAMs that are publicly available with coverage of both open and close sourced models. Due to budget, the models we tested in interactive evaluation are the six best-performing model settings (based on frequency of ranking as top five) which ensures a decent interactive capability for interactive evaluation: Qwen2-Audio \citep{chu2024qwen2}, DiVA-8B \citep{held2024distilling}, Typhoon-1.5 \citep{pipatanakul2023typhoon}, Gemini-1.5-pro \citep{team2024gemini}, GPT4o \citep{hurst2024gpt}, and ASR pipeline setting with whisper-large-v2 \citep{radford2022whisper} and Llama3-8B \citep{llama3modelcard}.

\paragraph{Metrics} For classification tasks, we report macro F1 scores to account for the importance of different classes due to class imbalance in some datasets. We compute PEDANTS score \citep{li2024pedants} for tasks requiring a short text response using the questions and reference answers. For ASR tasks, we report the Word Error Rate (WER).

\paragraph{Prompt}
Previous work \citep{wang2024audiobench} shows that some models like Qwen-Audio are prompt-sensitive. Therefore, we elicit models' responses using three different variations of text instruction prompts (see Appendix \ref{sec:prompt}). We take the average score of responses to different text prompts to get a more robust reflection of the models' capability.

\subsection{Model Performance on Static Benchmarks}

In general, close-sourced models generally top the leaderboard (Table \ref{tab:result}): GPT4o has the highest frequency of ranking first among all tested models (6 out of 14) and emerges as one of the top three for most tasks (11 out of 14). Gemini-1.5-pro also ranks among the top three models on more than half of the tasks tested (8 out of 14). It demonstrates strong performance in tasks related to speaker identity such as classification of accent (average F1 score of 24.5) and language (average F1 score of 68.8) as well as emotion classification tasks (average F1 score of 27.2).

Among the open-sourced models, Qwen2-Audio  and Typhoon-1.5 are the strongest performers based on the frequency of being among the top 3 models (Qwen2-Audio: 8/14; Typhoon-1.5: 7/14). Qwen2-Audio shows outstanding performance on gender (average F1 score of 69.1) and age classification (average F1 score of 12.3) which outperform all other models. Typhoon demonstrates best instruction following capability among all models, exceeding that of closed-models. 

We also perform an evaluation for the sequential ASR pipeline of Whisper-large-v2 and Llama3-8B-Instruct. It shows relatively good performance on benchmarks like CN-College-Listen (average F1 score of 62.6), IEMOCAP (average F1 score of 25.2), and MELD (average F1 score of 22.8), which means information in some of the data instances in those benchmarks can be inferred from textual content only. However, for every task there are end-to-end LAMs outperforming the ASR pipeline setting. This highlights that elements such as emotion, relationships, and sarcasm can be conveyed through vocal cues, necessitating \textbf{speech comprehension that goes beyond textual information}.

\subsection{Predictive Power of Static Benchmarks}
\begin{figure}[t]
\centering
  \includegraphics[width=\linewidth]{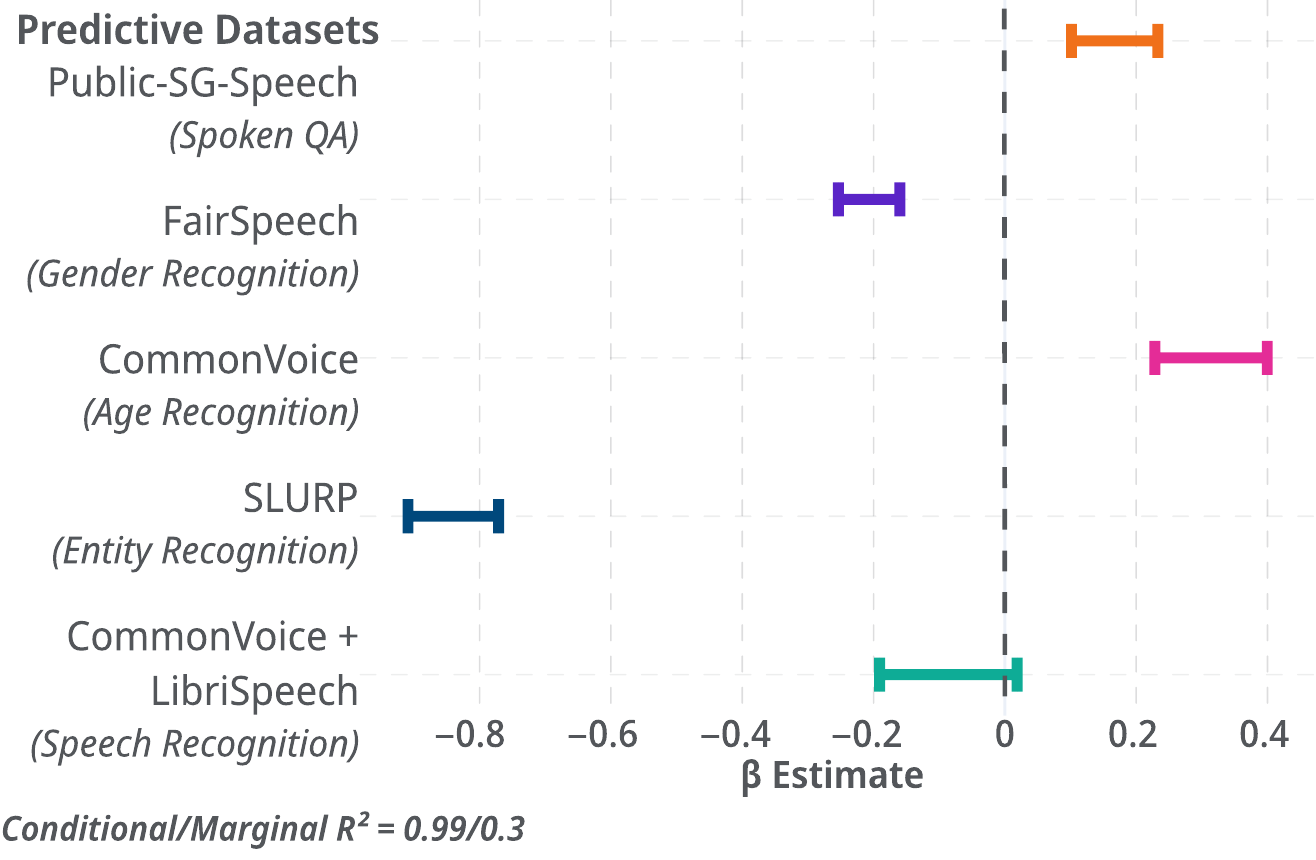}
  \caption{Mixed-effect regression of benchmark performance differences as a predictor of user preferences across models. 15 other features were pre-screened using VIF thresholding (threshold=10.0) with ties removed. Model fitting was performed fixed effects for benchmarks and random effects for model identity. The model achieved conditional/marginal $R^2$ of 0.99/0.30.}
  \label{fig:me_regression}
\end{figure}
The speed of iteration offered by static evaluations is invaluable during model development. As such, static benchmarks which correlate strongly with interactive evaluations, such as MMLU~\citep{hendrycks2021measuringmassivemultitasklanguage}, MTBench~\citep{zheng2023judgingllmasajudgemtbenchchatbot}, and AlpacaEval~\citep{dubois2024alpacafarmsimulationframeworkmethods}, are often used as proxies for text-based LLMs. Here, we test \textit{whether} this holds for speech evaluation and \textit{which} speech benchmarks are the best proxies.

First, using simple correlation checks, we find that no single metric strongly correlates with win rates in our evaluation ($\rho_s \leq 0.49$, $\tau \leq 0.33$). Furthermore, the model-capability matrix on static benchmarks is low-rank, with 95\% of the variance across our 20 tasks explained by five principle components. Similar to findings in \citet{ruan2024observational}, this suggests that despite a large number of benchmarks, only a few core axes of model performance are evaluated by these benchmarks.

While individual benchmarks are not clearly correlated, aggregated benchmarks could potentially have a smaller gap. To test this, we model head-to-head comparisons using logistic regression, where the dependent variable represents whether users preferred Model B over Model A. For each benchmark \( i \), we computed the normalized performance difference as \( \frac{s^i_B - s^i_A}{|s^i_A| + |s^i_B|} \), where \( s^i_A \) and \( s^i_B \) are the scores of Model A and Model B, respectively. To account for benchmark-specific effects while controlling for other underlying model features, such as output style, we use a mixed-effects model, treating benchmarks as fixed effects and model identity as a random effect. Our regression model gives a marginal \( R^2 \) of 0.3. This indicates that, unlike the text-only setting, current speech benchmarks have limited predictive power for user preferences.

However, some benchmarks did show significant positive correlations with user preferences. 
\textsl{CommonVoice - Age} showed the strongest positive association (\(\beta = 0.314\)). However, this correlation is particularly notable as all models performed below random chance on this benchmark, suggesting it may be \textbf{capturing systematic biases in model behavior} rather than meaningful capability differences.
\textsl{Public-SG-Speech}, a speech question answering task, also shows a moderate positive effect (\(\beta = 0.167\)). Notably, this task is, by construction, solvable using solely textual transcripts of the input speech since the questions were created based on the transcripts. This aligns with our observations on the overall strength of the pipeline model and the range of tasks found in Section \ref{sec:clio}.

\section{Conclusion}
Speech as an input modality has clear advantages for users when interacting with AI. It offers faster communication speed and enables the use of paralinguistic information. For users to benefit from these advantages, model developers must evaluate LAMs in ways aligned with user preferences. To test benchmark alignment, we collect over 7500 user voice queries and preferences for six different model settings, allowing us to analyze the expected use cases of LAMs and the models that users prefer the most during their interactions. Our results suggest that future benchmarks should focus more on testing models' ability to interact for efficiency purposes like knowledge expansion and task execution. With users' free-text explanations, we also identify key dimensions LAMs could work towards for better interactive capability: users still value the \textit{pragmatic value} (e.g. helpfulness, level of details, accuracy) \citep{garza2021artificial} and \textit{degree of adaptation} (e.g. language appropriateness, human-likeness) \citep{zargham2022want}.

While this work establishes key findings for speech-in text-out interaction, speech-in \textit{speech}-out models are a natural next step for LAMs. Exploring how our insights extend to such rich, real-time audio interactions presents key challenges to the existing norm of pairwise preferences, but represents an exciting direction for future work.

\section*{Limitations}
The current platform we use to evaluate LAMs' performances only supports single-turn interactions, and users are paid to interact with models and contribute their votes. Furthermore, since crowdworkers were given minimal constraints to better reflect their expectations, our evaluation focuses only on top-of-mind use cases. We expect that usage is likely to shift through long-term interaction with LAM systems as users become more familiar with the models' capabilities and become more comfortable with interacting naturally and emotively. These biases influence our current model ranks and likely negatively influence the ranks of commercial LAMs such as GPT-4o and Gemini which are designed for long-term uses.

Similarly, while our data collection did not require users to only utilize English, we only recruited participants who live in the United States. Therefore, our evaluation primarily assesses English language capabilities. In particular, this punishes models which aim for multilingual support, such as Typhoon and the Qwen models, which respectively include Thai and Chinese training data and occasionally respond to users in those languages rather than in English. However, this may not influence the ranks of Gemini and GPT-4o, which are also multilingual LAMs.

Finally, our setting is restricted to speech-in text-out format. As our analysis highlights, much of the qualitative user feedback focuses on the text output style rather than the capabilities of content. LAMs that are tuned for speech-in speech-out interaction, which only describes GPT-4o in our evaluations, likely have output styles that are more biased towards preferences for speech outputs and are likely penalized for this in our model ranks. While this is a limitation, we also think this highlights that model developers should likely tune their models to understand style preferences dependent on output modality instead of using a uniform treatment.

\section*{Ethical Statement}
Interacting with speech models can be associated with some ethical considerations about privacy and security, as we will have access to users’ voice data, which, if mishandled, could lead to unauthorized surveillance or data breaches. This study has been approved by the Institutional Review Board (IRB) at the researchers’ institution, and we obtained participant consent with a standard institutional consent form to record their voices. We anonymously store the data by applying advanced noise-masking techniques to the audio recordings, effectively reducing the recognizability of voices and ensuring that individuals cannot be easily identified. We will release the processed data only upon request and only for research purposes, ensuring strict control over its distribution and use.

\bibliography{custom}

\appendix
\renewcommand{\thefigure}{A.\arabic{figure}}
\setcounter{figure}{0}
\renewcommand{\thetable}{A.\arabic{table}} 
\setcounter{table}{0} 

\section{Prompt}
\label{sec:prompt}

\subsection{Humor Detection}
\noindent\textbf{Prompt 1} \\
Is the audio humorous?\\ A. Yes\\ B. No \\
\{\textit{input audio}\}

\noindent\textbf{Prompt 2} \\
Respond whether the input is intended to be humorous. Answer with a simple yes or no. \\
\{\textit{input audio}\}

\noindent\textbf{Prompt 3} \\
Based on the audio, indicate if it is humorous. Please give your answer as yes or no. \\
\{\textit{input audio}\}

\subsection{Sarcasm Detection}
\noindent\textbf{Prompt 1} \\
Is the audio sarcastic?\\ A. Yes\\ B. No \\
\{\textit{input audio}\}

\noindent\textbf{Prompt 2} \\
Respond whether the input is intended to be sarcastic. Answer with a simple yes or no. \\
\{\textit{input audio}\}

\noindent\textbf{Prompt 3} \\
Based on the audio, indicate if it is sarcastic. Please give your answer as yes or no. \\
\{\textit{input audio}\}

\subsection{Intent Detection}
\noindent\textbf{Prompt 1} \\
What is the intent of the speaker?\\ A. query alarm\\ B. remove alarm \\ C. set alarm \\ D. turn down audio volume \\
\{\textit{input audio}\}

\noindent\textbf{Prompt 2} \\
Respond what intent the input exhibits. Answer with ``query alarm'', ``remove alarm'', ``set alarm'', or ``turn down audio volume''. \\
\{\textit{input audio}\}

\noindent\textbf{Prompt 3} \\
Based on the audio, indicate the intent of the speaker. Please give your answer as ``query alarm'', ``remove alarm'', ``set alarm'', or ``turn down audio volume''. \\
\{\textit{input audio}\}

\subsection{Emotion Recognition}
\noindent\textbf{Prompt 1} \\
What is the emotion state of the speaker?\\ A. Angry\\ B. Happy \\ C. Sad \\ D. Neutral \\
\{\textit{input audio}\}

\noindent\textbf{Prompt 2} \\
Respond in a single word what emotion the input exhibits. Answer with ``angry'', ``happy'', ``sad'', or ``neutral''. \\
\{\textit{input audio}\}

\noindent\textbf{Prompt 3} \\
Based on the audio, indicate the emotion of the speaker. Please give your answer as ``angry'', ``happy'', ``sad'', or ``neutral''. \\
\{\textit{input audio}\}

The prompts above are used for IEMOCAP dataset. For MELD dataset, we apply the same format with the only changes in emotion options.

\subsection{Age Classification}
\noindent\textbf{Prompt 1} \\
What is the age of the speaker?\\ A. 18-22\\ B. 23-30 \\ C. 31-45 \\ D. 46-65 \\
\{\textit{input audio}\}

\noindent\textbf{Prompt 2} \\
Respond the age of the speaker based on the input. Answer with ``18-22'', ``23-30'', ``31-45'', or ``46-65''. \\
\{\textit{input audio}\}

\noindent\textbf{Prompt 3} \\
Based on the audio, indicate the age of the speaker. Please give your answer as ``18-22'', ``23-30'', ``31-45'', or ``46-65''. \\
\{\textit{input audio}\}

The prompts above are used for FairSpeech dataset. For CommonVoice dataset, we apply the same format with the only changes in age options.

\subsection{Gender Classification}
\noindent\textbf{Prompt 1} \\
What is the gender of the speaker?\\ A. female\\ B. male \\
\{\textit{input audio}\}

\noindent\textbf{Prompt 2} \\
Respond the gender of the speaker based on the input. Answer with ``female'' or ``male''. \\
\{\textit{input audio}\}

\noindent\textbf{Prompt 3} \\
Based on the audio, indicate the gender of the speaker. Please give your answer as ``female'' or ``male''. \\
\{\textit{input audio}\}

\subsection{Relationship Classification}
\noindent\textbf{Prompt 1} \\
Is the relationship between the two speakers more likely to be friend or relative?\\ A. friend\\ B. relative \\
\{\textit{input audio}\}

\noindent\textbf{Prompt 2} \\
Respond what relationship the two speakers have based on the input. Answer with ``friend'' or ``relative''. \\
\{\textit{input audio}\}

\noindent\textbf{Prompt 3} \\
Based on the audio, indicate the relationship between the speakers. Please give your answer as ``friend'' or ``relative''. \\
\{\textit{input audio}\}

\subsection{Accent Classification}
\noindent\textbf{Prompt 1} \\
'What is the accent of the speaker?\\ A. Australian English\\ B. Canadian English \\ C. England English \\ D. India and South Asia (India, Pakistan, Sri Lanka) \\ E. United States English
\{\textit{input audio}\}

\noindent\textbf{Prompt 2} \\
Respond the accent of the speaker based on the input. Answer with ``Australian English'' , ``Canadian English'', ``England English'', ``South Asia (India, Pakistan, Sri Lanka)'', or ``United States English''. \\
\{\textit{input audio}\}

\noindent\textbf{Prompt 3} \\
Based on the audio, indicate the accent of the speaker. Please give your answer as ``Australian English'', ``Canadian English'', ``England English'', ``South Asia (India, Pakistan, Sri Lanka)'', or ``United States English''.\\
\{\textit{input audio}\}

The prompts above are used for FairSpeech dataset. For CommonVoice dataset, we apply the same format with the only changes in age options.

\subsection{Instruction Following}
Please follow the instruction in the speech. \\ 
\{\textit{input audio}\}

\noindent\textbf{Prompt 2} \\
Respond to the instruction in the given audio. \\
\{\textit{input audio}\}

\noindent\textbf{Prompt 3} \\
Based on the audio instruction, please provide a response following it. \\
\{\textit{input audio}\}

\subsection{Speech Grounding, Entity Recognition, Language Classification, Question Answering}
In the datasets for these tasks, there are \textit{question} and \textit{options} provided so we structured our prompts as the following:

\noindent\textbf{Prompt 1} \\
\{\textit{question}\}\\ \{\textit{Option A}\}\\ \{\textit{Option B}\} \\ \{\textit{Option C}\} \\ \{\textit{Option D}\} \\
\{\textit{input audio}\}

\noindent\textbf{Prompt 2} \\
Respond to the question: \{\textit{question}\}. Answer with ``\{\textit{Option A}\}'', ``\{\textit{Option B}\}'', ``\{\textit{Option C}\}'', or ``\{\textit{Option D}\}''. \\
\{\textit{input audio}\}

\noindent\textbf{Prompt 3} \\
Based on the audio, \{\textit{question}\}. Please give your answer as ``\{\textit{Option A}\}'', ``\{\textit{Option B}\}'', ``\{\textit{Option C}\}'', or ``\{\textit{Option D}\}''. \\
\{\textit{input audio}\}

The prompts above are used for FairSpeech dataset. For CommonVoice dataset, we apply the same format with the only changes in age options.

\section{Correlation among Static Benchmarks}
\begin{figure*}[t]
  \includegraphics[width=\textwidth]{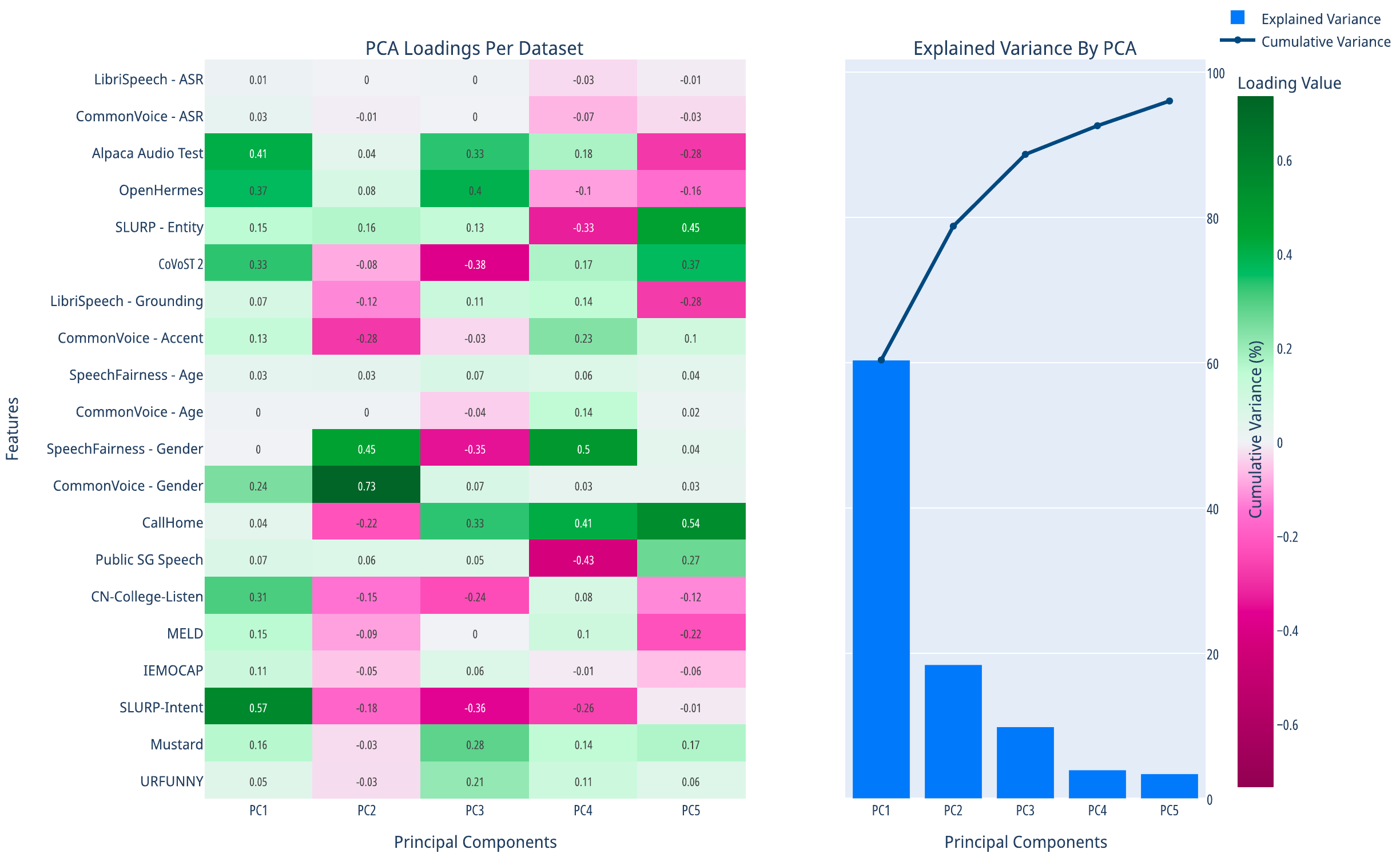}
  \caption{PCA Analysis of  model performance on 20 static benchmarks.}
  \label{fig:heatmap}
\end{figure*}

We compute correlation of LAM performance on 20 static benchmarks (see Figure \ref{fig:heatmap}). During regression analysis, we take the average model performance for benchmarks with a very high correlation as independent variables to remove high multicollinearity.

\section{Interactive Evaluation Interface}
\begin{figure*}[t]
  \includegraphics[width=\textwidth, trim=0.25in 5.5in 0.25in 0.25in, clip]{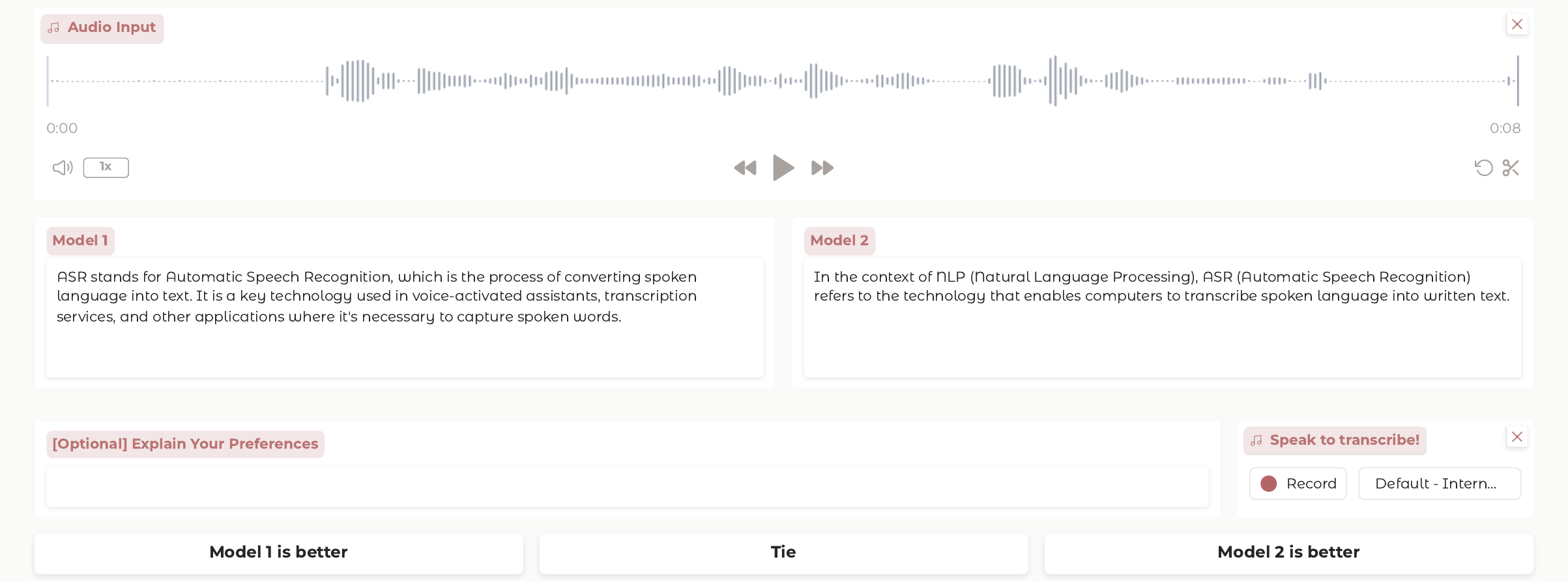}
  \caption{Gradio interface of interactive evaluation. Users record their own speech and audio without constraints and receive responses from two LAM systems anonymously. They then provide a binary preference between the models, and are provided the option to provide qualitative feedback through either voice or text.}
  \label{fig:interface}
\end{figure*}

In Figure \ref{fig:interface}, we illustrate the gradio interface we used to collect user preferences. Users can submit their voice input, get responses from two randomly sampled LAMs, contribute their votes and reasons for their votes.

\section{Model Ranking in Static and Interactive Evaluation}
\begin{figure*}[t]
  \includegraphics[width=\textwidth]{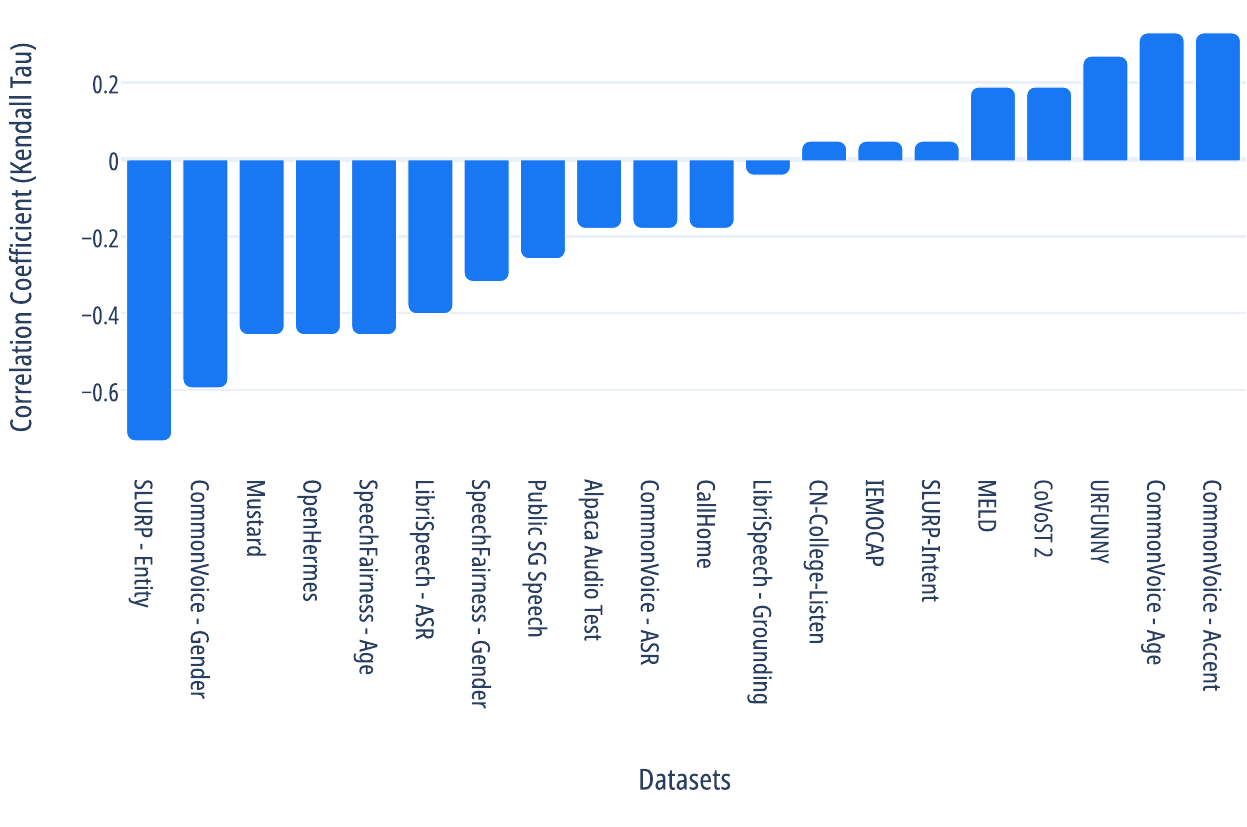
  }
  \caption{Kendall tau rank distance between model rank in interactive evaluation and that on different static benchmarks.}
  \label{fig:kendall_tau}
\end{figure*}

To understand the extent to which previous static benchmarks can reflect the relative interactive capability of LAMs, we also compare the result in interactive evaluation to that in static evaluation by computing the top-k Kendall Tau Distance \citep{bogomolov2024long} between the model rankings (see Figure \ref{fig:kendall_tau}).

We found that \textit{none of the 20 static benchmarks reflects exactly the same model rank in the interactive evaluation} with non-zero rank distance, suggesting that any single static benchmark is inadequate in reflecting the relative interactive capabilities of audio models and an interactive way of evaluation is essential. Among the 20 static benchmarks we tested, model ranks on Commonvoice age classification (rank distance: 0.20) is most similar to that in the interactive evaluation. On the other hand, model rank on SLURP speech entity recognition task (rank distance: 0.56) is most uncorrelated with that reflected in user preferences. The result is also reflected in our regression analysis.

\section{Logistic Regression}
\begin{figure}[h!]
\centering
  \includegraphics[width=\linewidth]{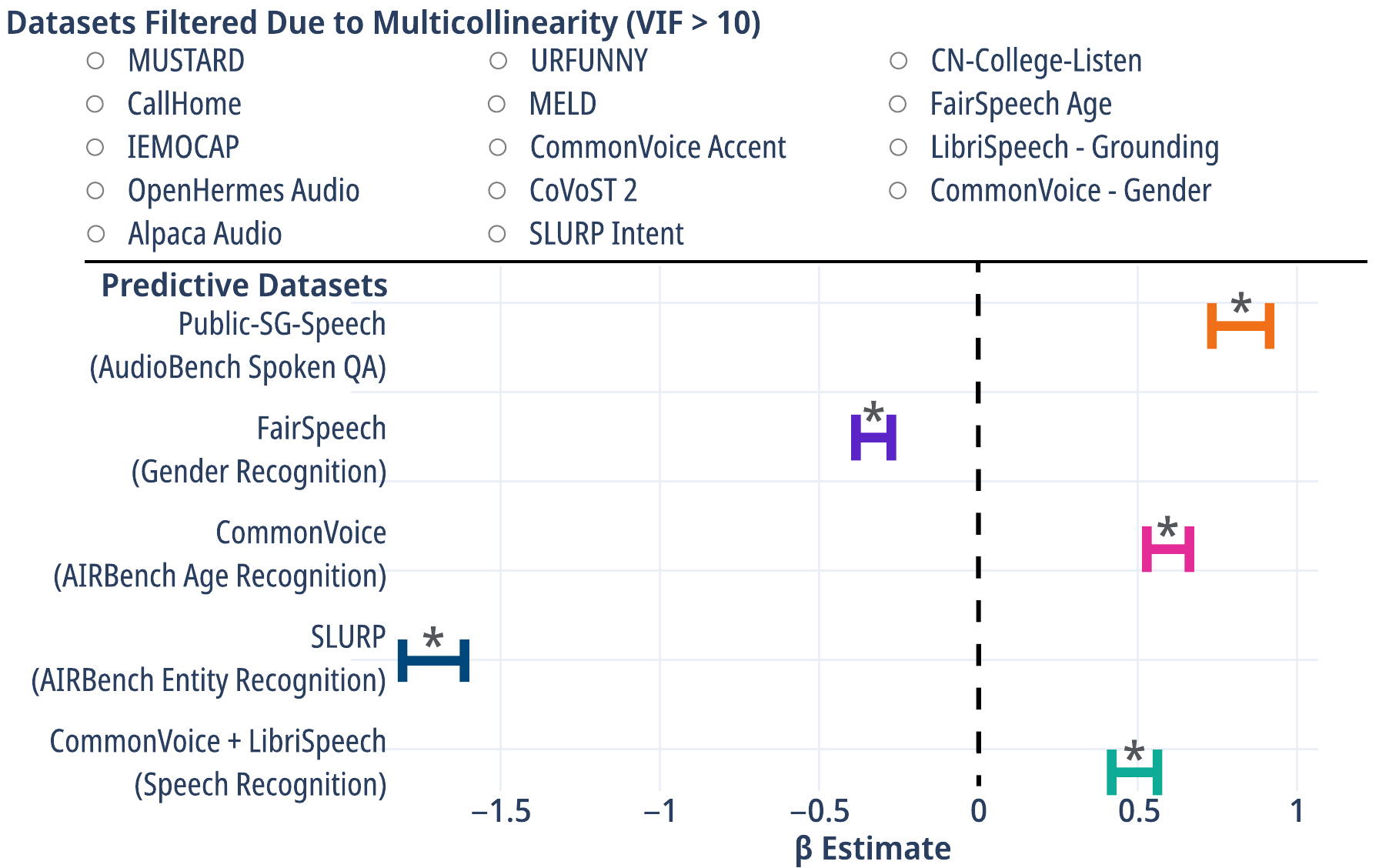}
  \caption{Forest plot demonstrating the effect sizes of static benchmarks in a logistic regression without mixed-effects for predicting individual user preferences. The results are overall consistent with our mixed effects regression with the only shift being the $\beta$ for ASR tasks. }
  \label{fig:logistic_regression}
\end{figure}

Besides mixed effect regression, we also perform a regular logistic regression without mixed effects to test the predictive power of static benchmarks with regard to interactive capability (Figure \ref{fig:logistic_regression}). We obtain similar findings as mixed effect regression where \textsl{Public-SG-Speech} and \textsl{CommonVoice Age Recognition} demonstrate positive effects. On the other hand, the effect of ASR benchmarks turns positive without taking mixed effects into consideration.

\end{document}